%% file: paper.tex
\RequirePackage{fix-cm}
\documentclass[smallextended]{svjour3}       
\smartqed  

\usepackage{booktabs} 
\usepackage[numbers,sort&compress]{natbib}

\DeclareMathAlphabet{\mathcal}{OMS}{cmsy}{m}{n}  
\usepackage{amsbsy,amsfonts,amsmath,amssymb,bbm,calc,color,dsfont,graphics}
\usepackage{graphicx,ifthen,latexsym,mathptmx,mathrsfs}
\usepackage{overpic,pifont,psfrag,rotating,stmaryrd,theorem}
\usepackage{bm}
\usepackage{float}
\usepackage{multirow}
\usepackage{multicol}
\usepackage[colorlinks,linkcolor=blue,citecolor=blue]{hyperref}
\usepackage{epstopdf}                   
\usepackage{geometry}                  
\spnewtheorem{method}{Method}{\bf}{\it}

\usepackage[misc]{ifsym}

\numberwithin{equation}{section}                        

\numberwithin{theorem}{section}                        

\numberwithin{lemma}{section}                           

\numberwithin{corollary}{section}                       

\usepackage{graphicx}

\journalname{}
 \usepackage{indentfirst}
\usepackage{algorithm}  
\usepackage{algorithmicx}  
\usepackage{algpseudocode}  
\usepackage{amsmath}  
\algdef{SE}[DOWHILE]{Do}{DoWhile}{\algorithmicdo}[1]{\algorithmicwhile\ #1} 

\usepackage{subfigure}  
\usepackage{float}  

\usepackage{siunitx}
\usepackage{amssymb}

\usepackage{makecell}

\usepackage{colortbl}

\usepackage{graphicx}
\usepackage{float} 
\usepackage{subfigure}
\renewcommand{\thesubfigure}{(\roman{subfigure})}
\makeatletter \renewcommand{\@thesubfigure}{\thesubfigure \space}
\renewcommand{\p@subfigure}{} \makeatother

\begin{document}
\title{     Solving the Travelling Thief Problem based on Item Selection Weight and Reverse Order Allocation                                                          
}


\author{
        Lei Yang \textsuperscript{~\Letter}\and
        Zitong Zhang \textsuperscript{}\and
        Xiaotian Jia \textsuperscript{}\and
        Peipei Kang \textsuperscript{}\and
        Wensheng Zhang \textsuperscript{}\and
        Dongya Wang \textsuperscript{}
}


\institute{
          Lei Yang 
           \at College of Mathematics and Information, South China Agricultural University, Guangzhou 510642, China \\
           \email yanglei\_s@scau.edu.cn
  \and
          Zitong Zhang 
           \at College of Mathematics and Information, South China Agricultural University, Guangzhou 510642, China
    \and
          Xiaotian Jia 
          \at College of Mathematics and Information, South China Agricultural University, Guangzhou 510642, China
    \and
          Peipei Kang  
          \at College of Computer, Guangdong University of Technology, Guangzhou 510006, China
    \and
          Wensheng Zhang  
          \at Institute of Automation, Chinese Academy of Sciences, Beijing 100190, China
    \and
          Dongya Wang 
          \at College of Engineering, Mathematics and Physical Sciences, University of Exeter, Exeter EX4 4QF, UK
}

\date{Received: date / Accepted: date}

\maketitle

\begin{abstract}

The Travelling Thief Problem (TTP) is a challenging combinatorial optimization problem that attracts many scholars. The TTP interconnects two well-known NP-hard problems: the Travelling Salesman Problem (TSP) and the 0-1 Knapsack Problem (KP). Increasingly algorithms have been proposed for solving this novel problem that combines two interdependent sub-problems. In this paper,TTP is investigated theoretically and empirically.An algorithm based on the score value calculated by our proposed formulation in picking items and sorting items in the reverse order in the light of the scoring value is proposed to solve the problem. Different approaches for solving the TTP are compared and analyzed; the experimental investigations suggest that our proposed approach is very efficient in meeting or beating current state-of-the-art heuristic solutions on a comprehensive set of benchmark TTP instances. 

\end{abstract}



%
%

\maketitle 
\input{paperbody} 

\end{document}

%% file: paperbody.tex
\section{Introduction}
\renewcommand{\labelitemi}{$\bullet$}
\label{intro}
Numerous practical applications include two or more sub-problems, many of which can be summarized as a combinatorial optimization problem. The combinatorial optimization problem is one of the most challenging problems. A combinatorial optimization problem usually involves traversing a search space in order to find an optimal solution or approximately optimal solution from a bounded solution set while maximizing (or minimizing) the objective function. Many interdependent components make it difficult to solve such problems: solving each component optimally not ensure obtaining an optimal solution to the overall problem.This type of problem is prevalent in supply chain management(like distribution,scheduling,loading,transporation,etc.)\cite{RN31,RN61},vehicle routing problem,logistics problem,etc.the reason why some optimization problems are difficult to tackle is that it is stated that interdependency among components in operational/dynamic problems play a key role in the complexity of the problems\cite{michalewicz2012quo}.\par
The TTP  combines two combinatorial problems, namely Travelling Salesman
Problem (TSP) and 0–1 Knapsack Problem (KP)
In order to demonstrate the complexity that arises by interdependency in multi-component
problems, a benchmark problem called Traveling Thief Problem
(TTP) was introduced by Bonyadi et al.\cite{RN23} in 2013.This problem can be illustrated in the following way.\par

A thief gets a cyclic journey through $n$ cities, and using a picking plan, picks $m$ items into a rented knapsack with restrained capacity. As items are picked up
at each subsequent city to fill the knapsack, the total profit of item and weight in the knapsack increases. The heavier the knapsack gets, the slower the thief becomes, therefore increasing the entire travelling time and hence the renting cost. The overall goal of the TTP is to concurrently maximize the picked total profit of item and minimize the renting cost. TTP can be considered to represent many real-world logistics issues\cite{RN20}.\par

From the above statement, it is clear that the two components of the TTP interact with each other. When the weight of the knapsack increases, it affects the speed of the thief, thereby increasing the rental time of the knapsack. When the tour is reselected, the order of the items in the corresponding city also changes. This interdependent relationship between the two components makes problem complicated.\par

Since the TTP is introduced by Bonyadi et al\cite{RN23} in 2013 as the benchmark problem for solving the multi-component and interdependence problems. Many scholars have proposed corresponding algorithms to solve this problem successively.
Polyakovskiy et al. \cite{RN17} was the first to create many benchmark instances and propose several heuristic algorithms to solve the TTP. An initial cyclic tour sequence is generated for TSP component by using the Lin-Kernighan heuristic\cite{RN75} in their paper, then select items under a fixed route until the optimal solution is obtained. In their first method for solving TTP named Simple Heuristic (SH), items are selected based on the score value. They also proposed some iterative heuristics called Random Local Search (RLS) and (1+1) EA based on flipping the picked
items with a specific probability.\par

Bonyadi et al.\cite{RN22} proposed a heuristic method to tackle
the TTP. In their approach, the TTP is disintegrated into sub-problems (TSP and KP). They process the two sub-problems while maintaining communication between them and then composed the solutions to get the final solution called CoSolver. They also proposed an approach is called density-based heuristics (DH) in their paper, a tour is generated for TSP, then a solution for KP is generated in a fixed tour.\par

Mei et al.\cite{RN48} introduced two evolutionary heuristic approach for
solving TTP. The first one is Cooperative Co-evolution (CC) which is to solve each sub-problem independently without considering the dependencies. The second one is the Memetic
Algorithm (MA) that solves this problem as a whole and considers the dependencies between each sub-problem.
An efficient Memetic Algorithm (MA) with the two-stage local search named MATLS is proposed by Mei et al.\cite{RN20} to solve the large scale TTP with several complexity reduction methods.\par

 In the KP of the TTP problem, a optimized picking plan called PackIterative was proposed by Faulkner et al.\cite{RN19} To avoid the bias toward the KP component, they proposed an insertion operator to optimize the tour iteratively for a fixed picking plan generated by the Lin-Kernighan heuristics\cite{RN75}. Several simple iterative heuristics (S1-S5) and some complex heuristics are proposed. According to the performance analysis, an simple iterative heuristic named S5 is the best performance on average among all the approaches. \par
 
 Yafrani and Ahiod \cite{RN21} introduced two heuristic algorithms. They compared two traditional types of search heuristics:population-based heuristic and single-solution heuristic.
The first method is a Memetic
Algorithm, called MA2B that uses 2-OPT operator and bit-flip operator. This method uses a genetic algorithm based on population evolution. The other approach is a single solution based heuristic method, named CS2SA which apply 2-OPT steepest ascent hill climbing heuristic and an adapted Simulated Annealing (SA) for efficient item picking to solve TTP. These two algorithms perform more competitive
compared to MATLS and S5 on many TTP instances.\par

Wagner \cite{wagner2016stealing} studyed a swarm intelligence approach, an idea of focusing on short TSP tours and good TTP tours for
solving the TSP component of the TTP based on ACO (Ant Colony
Optimization). This method is effective and computationally efficient for the small instance
of the TTP. However, its performance deteriorates
significantly for many large instances.
Neumann et al.\cite{neumann2018fully} investigated the underlying non-linear Packing While Traveling Problem (PWTP) of the TTP where the item are selected for a fixed route. they give an exact dynamic programming approach and a fully polynomial time approximation scheme (FPTAS) to solve this problem while maximizing the benefit.\par

Yafrani and Ahiod \cite{RN10} proposed two simple iterative neighborhood algorithms which are based on local search. The first approach named JNB (standing for Joint N1-BF) is a neighborhood-based heuristic that combines the N1 neighborhood (swapping two adjacent cities) of TSP and one-bit-flip of KP. The second one is named J2B (Joint
2OPT-BF) which is a combination of 2-OPT heuristic and one bit-flip heuristic.
Martins et al.\cite{martins2017hseda} introduced an approach named Heuristic Selection based on Estimation of Distribution Algorithm (HSEDA). This method applies the EDA probabilistic model using an approximation function to finding better heuristics for solving the TTP. Martins et al. confirmed this approach outperforms the other algorithms on most of the medium-sized TTP instances.\par

Wu et al.\cite{wu2017exact} proposed three exact algorithms and a hybrid approach for solving the TTP. They are Dynamic Programming, Branch and Bound search and Constraint Programming, respectively. El Yafrani et al.\cite{RN21} proposed an approache based on ConSolver with 2-OPT and Simulated Annealing (CS2SA).  After that, CS2SA* and CS2SA-R are introduced based on CS2SA. CS2SA* is an implementation of CS2SA with instance-based parameter tuning. CS2SA-R uses random restarts when no improvements in the state of returning the so far best solution. \par

Alharbi et al. \cite{alharbi2018design}] introduced a modified Artificial Bees Colony (ABC)
algorithm based on swarm intelligence for solving the TTP in an interdependent manner. It is efficient in mid-sized TTP instances compared to the state-of-the-art approaches.
Namazi et al.\cite{RN28,RN15} proposed an extended and modified form of the reversing heuristic to consider both the TSP and KP components concurrently. Items regarded as less profitable and selected in cities beginning in the reversed segment are substituted by items that tend to be equally or more profitable and not selected in the later cities.
Maity et al.\cite{maity2020efficient} introduced scoring value which is calculated by they proposed formulation to pick items for a fixed picking plan generated by the chained Lin-Kernighan heuristics.\par 

In this paper, We mainly focus on the KP component of the TTP, Because we believe the KP component of the TTP is more critical as compared to the TSP component for optimization. A near-optimal tour (TSP component of the TTP) is generated by Lin–Kernighan Heuristic (LKH). We chiefly discuss whether the weight of the items have a greater determinant of the final profit than other item attributes(location in the tour and value of item).Therefore, A formula for calculating the value of the item to calculate the impact of the item on the final profit is proposed. We believe that the final profit is not only related to the value, weight, and location of the items, but the order in which the items are picked up. The effect of the selection order of items on the final profit is also discussed in this article.\par

The rest of this paper is organized as follows. In Section 2, a background about the TTP and some heuristics are introduced, In Section 3, The proposed algorithms are applied on some TTP instances and the experimental results are reported and discussed in Section 4. Finally, Section 5 concludes the paper and outlines some future directions.

\section{Background}
In this section,we present brief background introduction and formulation about the TTP. Some common heuristics for TSP and KP are briefly revisited.\par

\subsection{The Travelling Thief Problem}
The Travelling Thief Problem is a combination of two well known benchmark problems,namly the Travelling Saleman Problem (TSP) and the Knapsack Problem (KP). In the TTP, we consider $n$ cities and the associated distance matrix $\{d_{i,i'}\}(i\neq i')$,The distance is the distance between each pair of cities, $d_{i,i'} = d_{i',i}$ ($i,i' \in \{0,\cdots,n\}$ ). There are $m$ items scattered in these cities. Each item $j (j \in \{0,\cdots,m\})$ is located at city $l_j$ having a profit $p_j >0$ and a weight $w_j >0$. A thief starts from the first city to visit all these cities only once and pick up a subset of the items 
available in each city. We suppose each item is avaliable in only one city and we note $ A_i \in \{1,\cdots,n\}$ as the avaliablity vector that contains the reference of the city where the item $i$ is located.
The cyclic tour is designed by using a permutation of n cities. Given a tour $c$,We define $c_k = i$ as $i$ is the $k$-th city in the tour $c$,and we define $c(i) = k$ as the location of city $i$ in the tour $c$ is $k$.
A knapsack with a maximum weight capacity W and a rent rate R per time unit is rented by
the thief to carry the picked items.
$W$ denotes the maximum capacity of the knapsack, $v_{min}$ (when the knapsack is empty) and $v_{max}$ (When the knapsack is full of items) are the minimum and maximum possible velocity, respectively. The total weight of the items in the knapsack must not exceed the maximum weight limit. The speed of the thief varies with the weight of the backpack. The heavier the knapsack gets, the slower the thief becomes in the tour. \par

A solution of the TTP is represented as follows:
\begin{itemize}
    \item The tour $c = (c_1,c_2,\cdots,c_n)$ is a vector containing the permutation of cities.
    \item The picking plain $z = (z_1,z_2,\cdots,z_m)$ is a binary vector determined that item is picked if $z_j = 1$,or not picked if $z_j = 0$.
\end{itemize}

The interdependence of the two sub-problems in the TTP problem is reflected in the dependence of the  speed of the thief and the total weight of the knapsack.The total weight of the items picked from city $i$ is given in equation \ref{eq:weightInCity}, and the total weight of the items picked from the begin city to the $k$-th city in the cyclic tour $c$ is given in equation \ref{eq:weightAll}. The velocity of the thief decreases linearly with the increase of the total weight of the knapsack. We note $v_{c,z}(k)$  as the velocity at the city $c_k$ in equation \ref{eq:speed}, and note $C = \frac{v_{max} - v_{min}}{W}$.
	\begin{equation}
		W_z(i) =  \sum_{l_j=i}^{} w_j z_j   \label{eq:weightInCity}
	\end{equation}
	\begin{equation}
	W_{c,z}(k) =  \sum_{k'=1}^{n} W_z(c_{k'})    \label{eq:weightAll}
	\end{equation}
	\begin{equation}
	v_{c,z}(k) = v_{max} - W_{c,z}(k) \times  C\label{eq:speed}
	\end{equation}
	
The goal of TTP is to find out a proper tour $c$ and a picking plan $z$ to maximise the total gain G(c,z) defined in equation \ref{eq:Gain}. In other words, the goal is to maximise the total profit while minimise the total renting cost of knapsack. The total weight of the picked item must not exceed the capacity of the knapsack.

\begin{equation}
	maximization \quad G(c,z) = \sum_{i=1}^{m} p_i z_i - R \times T(c,z) \label{eq:Gain}
\end{equation}

\begin{equation}
     T(c,z) = T_{c,z}(n+1) = T_{c,z}(n) + \frac{d(c_n,c_1)}{v_{c,z}(n)}
\end{equation}

\begin{equation}
     T_{c,p}(k) = \frac{\sum_{k'=1}^{k-1} d(c_{k'},c_{k'+1})}{v_{c,z}(k')}
\end{equation}

\begin{equation}
    s.t. : \sum_{k=1}^{m} w_k z_k \le W
\end{equation}

\subsection{In the TSP component}
Lin–Kernighan Heuristic (LKH) introduced by Lin and Kernighan \cite{RN75} is a generalization of the 2-OPT search algorithm for solving TSP.This algorithm and the Chained Lin–Kernighan heuristic (CLKH) often are often used to optimize TSP problems and to initialize TSP component of the TTP. To solveing the TSP component of the TTP,2-OPT (a segment reversing heuristic) is often used for modifying the tour $c$. On a tour given by two position $i$ and $j$ $(1 < i < j \le n)$, the order of the visited cities between these two position is reversed to get a new tour. The 2-opt function is defined as follows.


\begin{gather}
    c'(j-k,i+k) = 2OPT(c(i+k,j-k)) \label{eq:2opt} \\
     s.t. : 0 < i <j \le n ;  \quad 0 \le k \le j-i \notag
\end{gather}

The Delaunay triangulation method \cite{delaunay1934sphere} is used as a candidate generator for the 2-OPT heuristic. Generated candidate objects by the Delaunay triangulation can reduce the time complexity without significantly reducing the quality of the solution.Besides,by tracking time and weight information at rach city of a given tour in the TTP also can reduce the total time budget.

\subsection{In the KP component}

For solving the KP component, the bit-flip operator introduced by Faulkner et al. \cite{RN19} is often used for optimizing the packing plan $z$.
The bit-flip operator works iteratively by flipping
each bit in the picking plan $z$. Given a picking plan $z$ and a selected item j, the picking state $z_j$ is flipped from 0 to 1 or vice versa to obtain a new picking plan $z'$.If the performance is improved after bit-flip operation, we will keep this state, otherwise we will continue the bit-flip operation until the termination condition is reached.

\section{Proposed approach}

This section describes our idea of optimizing TTP, illustrates it with examples, and proposes a algorithm for solving TTP.\par

 In the definition of the TTP given in section 2, a tour and an item picking plan are required. First, the TSP search heuristic render TTP solution with an empty tour. Then, the items are required to insert into the empty tour to increase the profit. It is common to employ a proper measure of elements of a problem to make a judgment. A scoring function for picking items is introduced to decide which item should be picked. This function is commonly based on profit, weight, and distance from the city where the item is picked to the end city. Generally, the function is $ScoreValue_{i,k}( c ) = \frac{p_k}{w_k \times \sum{}{}d_{i,1}}, $ (or other form ). where $c$ is the tour, $p_k$ is the profit, $ w_k$ is the weight of the item $k$ in city $i$, $\sum{}{}d_{i,1}$ is the distance from the city $i$ where the item k is picked to the end of the tour, $i \in (2 ,\cdots,n), k \in (1, \cdots, m).$ The higher the score of an item, the more likely it is to be picked up. However, If an item is of high profit but very heavy and is close to the start city in the tour, the item’s score value is also relatively high, according to the principle that the greater the score value of the item, the more likely it is to pick up the item. Picking up this item will result in other items in the tour that are close to the end city, and items with high score value may not be picked up. In this case, it may slow down the speed of the thief, spend more time, and make the total profit smaller.\par
 
 Without considering picking up other items, the change of profit caused by inserting an item $k$ in city $i$ to the overall profit is $\Delta p'_{i,k} = p_k - R \times \frac{\sum{}{}d_{i,1}}{v_{max} - w_k \times C} $, However, the fact that a single item changes the overall profit is closely related to the items previously selected. The actural change of profit caused by inserting an item $k$ in city $i$ is  $\Delta p'_{i,k} = p_k - R \times \frac{\sum{}{}d_{i,1}}{v_{max} - W_{c,z}(k) \times C} $ we noted. $\Delta p'_{i,k}$ may be a positive number ($\Delta p'_{i,k} > 0 $), Due to the accumulated weight of the items picked up before,  $\Delta p_{i,k}$ may become a negative number ($\Delta p_{i,k} < 0$). This means that the impact of a single item on the overall profit needs to take into account the cumulative effect of the items picked up previously.\par
 
 Besides, owing to the cumulative effect of the weight of the picked items, the order in which items are picked up needs to be taken into account. In the following section, we use a function on sequence of numbers or sets called \emph{Reverse-search}, For any position $k$ of given sequence of $n$ number $S = (s_1,s_2,\cdots,s_n)$, the \emph{Reverse-search} function is defined as follows.

\begin{gather}
    S'(s_{n}, \cdots, s_{1}) = Rev(S(s_1,\cdots,s_n))  \label{eq:Rev}
\end{gather}
This function defined in equation \ref{eq:Rev} is similar to the 2OPT function; the difference is that the elements in the 2OPT function are numbers, and the elements in this function can be numbers or sets. For TTP, the elements in this function are the sets of item attributes (weight, value, scoring value, etc.) in the city.
 \par
 
\begin{figure}[h]
	\centering
	\includegraphics[scale=0.75]{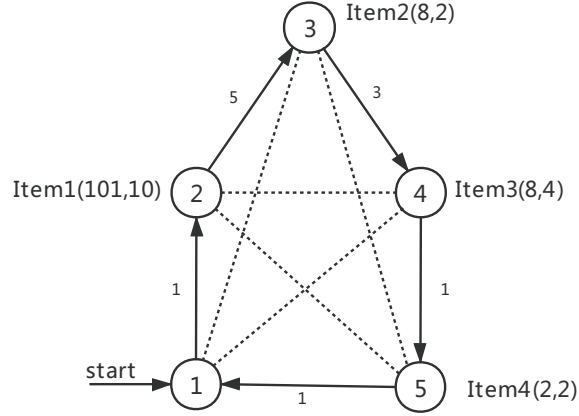}
	\caption{An example TTP instance}
	\label{fig:TTP_example}
\end{figure}

 To further explaination, the following example can be illustrated.As an example, consider the simple TTP instance shown in Figure \ref{fig:TTP_example}, which illustrates an example of the TTP with $n = 5$ cities and $m = 4$ items. Each city is assigned
with a set of items except the first city (No items in the starting city),The nodes represents the citys. For example, node 2 is
associated with the item of profit $p_2 = 101$ and weight $w_2 = 10$. Suppose that the capacity of the knapsack $W = 10$, renting rate $R = 1$, maximun speed $v_{max} = 1$ and minimum speed $v_{min} = 0.1$, Furthermore, assuming that an interin solution has the tour $c = [1, 2, 3, 4, 5]$ and the picking plan $z = [1, 0, 0, 0]$ (If the item picking plan is based on the score function mentioned above $ScoreValue_{i,k}( c ) = \frac{p_k}{w_k \times \sum{}{}d_{i,1}},$ then the score value of the items are $s_1 = 1.01$, $s_2 = 0.8$, $s_3 = 1$, and $s_4 = 1$, According to the principle of picking up high-scoring items first and the current total weight $W_c$ of items picked up must not exceed the maximum capacity $W$), Since in the subset of the tour 1-2 no item is picked up,
hence the thief travelled with maximum velocity $v_max = 1$, From city 2,item2 is picked which makes the current speed $v_c = 0.1$ and the knapsack is full. The optimal objective value will be $G(c,z) = 101 - 1 \times (1 + \frac{10}{0.1}) = 0$, Assuming that the tour is fixed and the picking plan $z' = (0, 1, 1, 1)$. Travel time from city 1 to 3 is $1+5
 = 6$. From city 3, item2 is picked
which makes $v_c = 0.82,W_c = 1$. Thus, the travel time from city
3 to 4 is $\frac{3}{0.82} = 3.66$. From city 4 item3 is picked which makes
$v_c = 0.46,W_c = 6$. The travel time from city
4 to 5 is $\frac{1}{0.46} = 2.17$. From city 5 item4 is picked which makes
$v_c = 0.28,W_c = 8$. The travel time from city
5 to 1 is $\frac{1}{0.28} = 3.57$.
Therefore, $T (c, z') = 6 + 3.66 + 2.17 + 3.57 = 15.40$, and the objective
value $G(x, z') = 18 - 1 \times (15.40) = 2.60$.\par

\subsection{EW formula}
Based on the above inspiration, we proposed a new selecting items approach. and choosing potential items in reverse order according a specific formula proposed by us. Our motivation for this approach is twofold: (i) Due to the cumulative effect, the weight of the item has a greater impact on the final profit than other item attributes (value, location, etc.). (ii) Prioritizing the selection of high-value items at the end of the travel route can maximize the final profit.\par

The formula for expanding the effect of item weight is proposed (EW formula). A new scoring function based on location, weight, and profit of the item is introduced to generate a score for the item $k$ placed in city $i$ as follows:
\begin{equation}
    Score_{i,k}( c ) = \frac{p_k}{w_k^{\alpha} \times \sum{}{}d_{i,1}} \label{eq:myScore}
\end{equation}

where $\sum{}{}d_{i,1}$is the distance from city $i$ to the end of a given tour $c$, $p_k$ is the profit and $w_k$ is the weight of the item $k$. we proposed exponents applied to the weight of an item to manage the impact on the score value. Our preliminary study shows that keeping the exponent on the weight of an item results in better objective value on large-scale instances. To get the best performing values of $\alpha$, we perform an experimental study over dozens of times to calculate the objective value. The best objective value is found when the value of $\alpha$ = $1.5$, where $\alpha > 1$.\par

\begin{figure}[h]
	\centering
	\includegraphics[scale=0.55]{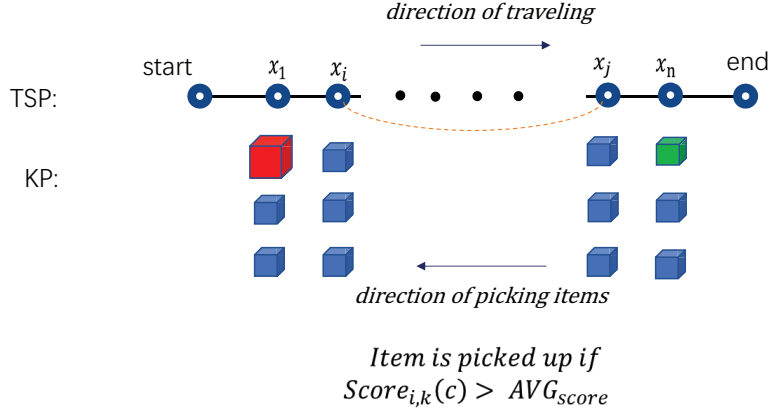}
	\caption{Illustrative diagram of proposed method}
	\label{fig:reversePic}
\end{figure}

\subsection{Reverse order searching approach}

The constructing process starts by calculating $Score_{i,k}$ for each item according to formula introduced in equation \ref{eq:myScore}, the items are sorted
according to the non-decreasing order of $Score_{i,k}$ . We will discard all items with $Score_{i,k} < 0$ and keep the items with a $Score_{i,k} > 0$. We suppose there are a total of $l$ items ($l \le m$). We calculate the ratio of the number of picked items $l$ to the total number of items $m$ as $r$ ($r = \frac{l}{m}$). Take the average value of all items with a $Score_{i,k} > 0$ as $AVG_{score}$. The function can be summarized as follows.
\begin{equation}
    AVG_{score}( c ) = \frac{\sum_{i=1}^{l}Score_{i,k}}{l} \label{eq:myScore}
\end{equation}

For every item $k$, if $Score_{i,k} > AVG_{score}$, then the item $k$ is a potential one. In city $i$, the potential items are denoted as the set $ s_i$, where $s_i$ contains zero, one or more items. The high-value items is picked up in reverse order along the given tour. If inserting item $k$ does not decrease the objective value
and also fits into the knapsack then items $k$ is picked, otherwise, process
the next item, and so on. For all items on the entire route, we mark the picked items as $Above\_AVG(s_1, \cdots, s_n)$. Then, picking up items in reverse order until the item is picked up to $r$ of the knapsack capacity. The proposed approach we noted as RWS (Based on Item Selection Weight and Reverse Order Allocation). In order to facilitate readers to follow 
our algorithm, the illustrative diagram of proposed method shown in Figure \ref{fig:reversePic}. In the entire travel tour, high-value and low-weight items in cities near the end (green item in the picture) are more likely to be selected than high-value and low-weight items in cities near the beginning (red item in the picture).
\par

\begin{algorithm}[h]
  \caption{ Algorithm framework}
  \label{alg:frame}
 \begin{algorithmic}[1]

\State $(c^*, z^*) \gets \emptyset$ \{best\ solution\}
\State Set the current picking plan $ z = \emptyset$ and current weight $W_c =0$
\State Set current tour $ c = \emptyset $, and calculate $G(c,z)$

    
        \While{not global timout} 
            \State $c \gets$ $LKTour()$
            \State $z \gets InitPickingPlan(c)$
            \State $(c,z) \gets $ TSPSolver$(c, z)$
            \State $(c,z) \gets $ KPSolver$(c, z)$
            \If{$G(c,z) > G(c^*,z^*)$}
                \State $ (c^*,z^*) \gets (c,z)$
            \EndIf
        \EndWhile
    
    \State \Return{$(c^*,z^*)$}
    \label{code:recentEnd}
  \end{algorithmic}
\end{algorithm}

 \begin{algorithm}[htb]
  \caption{ Initial Picking Plan}
  \label{alg:init}
  \begin{algorithmic}[1]
    
\State Compute the score of the each item $I_k \in m $ by proposed formulation for the given tour $c$
\State Sorting the items of $m$ in descending order according to their score value
\State Calculate the value of items greater than 0 and calculate their average $AVG_{score}$ and max value $MAX_{score}$
\State Set current packing plan $z = \emptyset$ and current weight of knapsack $W_c = 0$
\State Set $\beta \in [0, 1]$, which is set according to the size of the instance 
        \While{$W_c  < W  $} 

                \For{$i\gets n$ To $2$ }
                 \If {$ Score_{i,k}  > AVG_{score} + (MAX_{score} - AVG_{score}) \times \beta $} 
                \State  add item $I_k$ to the picking plan $z = z \cup \{I_k\}$
                \State set $W_c = W_c + w_k$
                \EndIf
                 \If {$ W \times r \le W_c$}
                \State Break
                \EndIf
                \EndFor
                
                \For{$j\gets 1$ To $m$ }
                
                 \If {$ W \geq W_c$}
                \State Insertion heuristic
                \Else
                \State Break
                \EndIf
                \EndFor
                
            \If{$G(c,z) > G(c,z^*)$}
                \State $ z^* \gets z$
            \EndIf
        \EndWhile

                \State \Return{$z^*$}
  \end{algorithmic}
\end{algorithm}

The Algorithm \ref{alg:frame} describes the basic framework for solving TTP. Based on the above ideas, the initial picking plan is introduced in Algorithm \ref{alg:init}.
 The idea of Algorithm \ref{alg:init} can be explained as follows: At first,The new initial cyclic tour is generated by Lin-Kernighan heuristic, the priporities of the items are determined by the formula in equation \ref{eq:myScore}, The higher the score of the items, the higher the priority of the item being picked up. Then, pick out items with positive scores (Items with negative scores do not contribute to total profit) and calculate their average score $AVG_{score}$ and max score $MAX_{score}$. 
Afterwards, select items with a score greater than the average in the reverse order of the cities in the travel tour. The capacity constraint is imposed as a global constraint. That is, any insertion that result in the violation of the capacity constraint will be prohibited. The Insertion heuristic is based on YI Mei \cite{RN20} in the Algorithm \ref{alg:init}. Finally, restore the picking plan $z^{*} = z$. The $TSPSolver$ adopts 2-OPT heuristic search to optimize TSP component. Besides, the Delaunay triangulation is introduced as a candidate generator for 2-OPT heuristic. In the $KPSolver$,  both the BitFlip operator and the simulated annealing metaheuristic are commonly used algorithms for the KP component. In this article, we use the simulated annealing algorithm to solve the KP component.


\section{Experimental Study}
In this section, the experimental setup of the TTP is described and the comparative results are investigated with other state-of-the-art approaches. 
\subsection{Benchmark instances and experimental setup}
We use the comprehensive set of TTP instances for our investigations that from Polyakovskiy et al.\cite{RN17}. The two components of TTP are balanced in these instances in such a way that the near-optimal solution of one subproblem does not guarantee the optimal solution of annther one.\par

The TTP dataset introduces the following diversification parameters resilting in 9720 TTP instances, and these instances are generally based on the instances from TSPLIB by REinelt \cite{reinelt1991tsplib} and the types of the kanpsacks introducesd by Martello et al.\cite{martello1999dynamic}. We consider a subet of the TTP
library to perform our tests:
$\ eil76$,\ $kroA100$,\ $ch130$,\ $u159$,\ $a280$,\ $u574$,\ $
u724$,
\ $dsj1000,$\ $rl1304$,\ $fl1577$,\ $d2103$,\ $pcb3038$,\ $fnl4461$,\ $pla7397$,\ $rl11849$,\ $usa13509$,\ $brd14051$,\ $d15112$,\ $d18512$,\ $pla33810$.\par
These instances cover small, medium, and large size instances with differenct characteristics. We denote the 4 categories as A, B, C, and D (Category C and D have same KP type and Item factor):

\begin{itemize}
    \item Category A: 1 item in each city, item values and weights are bounded and strongly correlated, small capacity of the knapsack.
    \item Category B: 5 item in each city, uncorrelated KP but similar item weights, average capacity of the knapsack.
    \item Category C: 10 item in each city, uncorrelated KP, high  capacity of the knapsack.
    \item Category D: 9 item in each city, uncorrelated KP, high  capacity of the knapsack.
\end{itemize}

According to different data types A, B, C, D, the $\beta$ metioned in Algorithm \ref{alg:init} is 1, 0.65, 0.5, 0.5, respectively. The experiment setup is adopted in all experiments. All solvers are run on each TTP instance 10 times indepently, and all the algorithms have a maximum runtime limit of 600 seconds. All experimnts are performed on a computer with Intel(R) Core(TM) i5-8500 CPU(3.00GHz).\par

\subsection{Comparison of algorithms}
In order to gain further insights into the performance of each solver for solving the TTP, We performed a statistical analysis of Friedman's test \cite{zimmerman1993relative} for all methods.It is an alternative to repeated measures one-way analysis of variance \cite{cuevas2004anova}. It is a non-parametric test. When the dependent variable is ordinal, it is used to find the difference between groups and be used for continuous data. To measure the consistency of the results, the ralative standard deviation (RSD) is introduced. The formula is defined as $ RSD = \frac{S}{\overline{x}} \times 100\%$, where $S$ is the mean of the standard deviation, $\overline{x}$ is the arithmetic mean. For quality measures of all methods, we adopt average ranking and Friedman's test ranking, and calculate the ranking of each method based on the target value of each TTP instance for the average ranking method. And then, we compute the average ranking for each method.\par

In Friedman's test ranking, the formula of test statistic is defined as $F = \frac{12}{nk(k+1)} \sum r^2-3n(k+1)$, where $n$ donate the number of instances, $k$ is the mean of the number of methods. First, the ranking of each method is calculated for each instance. And next, we compute the sum of the rank (r) of each method. Then, the probability value (p) and the degrees of freedom (d)  are applied to calculate the critical chi-square value. The null hypothesis is rejected if the F value is greater than the critical chi-square value. Finally, the average rank of each method is calculated. The size of Friedman's 
test value can be used to measure the quality of the various algorithms listed.\par

\begin{table}[htbp]


\centering

\caption{Results for category A}

\label{Tab01}

\begin{tabular}{ccccccccc}

\toprule[0.3mm]

\multirow{2}{*}{\textbf{Insance}} & \multicolumn{2}{c}{$MATLS$} & \multicolumn{2}{c}{$S5$} & \multicolumn{2}{c}{$CS2SA^*$} & \multicolumn{2}{c}{$RWS$} \\

\cmidrule(r){2-3} \cmidrule(r){4-5} \cmidrule(r){6-7} \cmidrule(r){8-9}

&  $Mean$      &  $RSD$  
&  $Mean$      &  $RSD$  
&  $Mean$      &  $RSD$  
&  $Mean$      &  $RSD$    \\

\toprule[0.1mm]

$eil76 $             &3705(3)                          & 1.35                    & 3742(2)                   & 0                 & 3425(4)           & 0.31          & \textbf{3765}(1)           & 2.52                   \\
$kroA100 $            &\textbf{4660 }(1)                        & 1.36                   & 4283(4)                   & 0           & 4435(3)           & 1.07                 & 4445(2)           & 1.9                   \\
$ch130$               &8876(4)                         & 0.79                    & \textbf{9250}(1)                   & 0           & 8964(3)           & 0.63                  & 9013(2)           & 0.03                   \\
$u159$                &8403(4)                          & 1.40                   & \textbf{8634}(1)                  & 0           & 8452(3)          & 0                     & 8627(2)           & 0.33                   \\
$a280$              &17678(4)                          & 0.54                    & \textbf{18406}(1)                   &0.01           & 17728(3)           & 0.22              & 17743(2)           & 0.53                   \\
$u574$            &26121(3)                          & 2.30                    & \textbf{26957 }(1)                  & 0.10           & 26100(4)           & 0.03           & 26366(2)           & 0.6                    \\
$u724$               &48980(3)                          & 1.25                    & \textbf{50313 }(1)                  & 0.12           & 49623(2)           & 0.03          & 48794(4)           & 1.0                    \\
$dsj1000$                &143699(2)                         & 0                    & 137653(4)                   & 0.16           & \textbf{144219}(1)          & 0          & 140656(3)           & 0.7                    \\
$rl1304$              &75800(3)                          & 1.26                    & \textbf{80067 }(1)                  & 0.86           & 75825(2)           & 0.01          & 75206(4)           & 0.7                   \\
$fl1577$            &88375(3)                          & 0.41                    & \textbf{92328}(1)                   & 1.25           & 88259(4)           & 0.16            & 88923(2)           & 0.7                    \\
$d2103$               &113005(4)                          & 0.45                    & \textbf{120482   }(1)                & 0.2           & 118844(2)           & 0          & 118338(3)           & 0.05                     \\
$pcb3038$                &148265(3)                          & 1.18                    & \textbf{160006}(1)                   & 0.15          & 145837(4)          & 0.6          & 148973(2)           & 0.6                     \\
$fnl4461$              &247553(2)                          & 0.40                    &\textbf{ 262237}(1)                   & 0.11           & 239287(4)           & 0.42          & 241291(3)           & 0.3                   \\
$pla7397$            &365613(2)                          & 1.32                    & \textbf{395156 }(1)                  & 0.56           & 315153(4)           & 0          & 315386(3)           & 0.15                    \\
$rl11849$               &661392(2)                          & 0.29                    & \textbf{707183   }(1)                & 0.24           & 658519(3)           & 0.05          & 653857(4)           & 0.33                    \\
$usa13509$                &747885(2)                          & 0.53                    & \textbf{809623}(1)                   & 0.35           &683123(3)          & 0.2          & 677983(4)           & 0.66                    \\
$brd14051$              &815602(2)                          & 0.36                    & \textbf{875008}(1)                   & 0.25           & 800495(3)           & 0.06          & 798787(4)           & 0.12                  \\
$d15112$            &871153(2)                          & 0.52                    & \textbf{939726 }(1)                  & 0.48           & 870253(3)          & 0.12             & 868019(4)           & 0.27                    \\
$d18512$               &996582(2)                          & 0.84                    & \textbf{1072308}(1)                   & 0.21           & 964625(3)           & 0.23          & 962781(4)           & 0.32                     \\
$pla33810$                &1730352(4)                          & 0.92                    &\textbf{1870306}(1)                   & 0.62           & 1778256(3)          & 0.31         & 1781984(2)           & 0.36                    \\
\midrule

\multirow{1}{*}{\textbf{Average ranking}} & \multicolumn{2}{c}{$2.75$} & \multicolumn{2}{c}{$1.35$} & \multicolumn{2}{c}{$3.05$} & \multicolumn{2}{c}{$2.85$} \\



\bottomrule
\end{tabular}
\end{table}

The results of the comparison study between the proposed method and three other state-of-the-art algorithms are shown in Table 1-3. For each instance, 10 independent runs are performed. The best mean objective values are highlighted in bold, the mean objective value is regarded as the quality of the solution to compare the performance of the algorithms, and the results of the Friedman’s test-based ranking for each method are presented in the last row of each table.\par

\begin{table}[htbp]
\centering
\caption{Results for category B}

\label{Tab02}

\begin{tabular}{ccccccccc}

\toprule[0.3mm]

\multirow{2}{*}{\textbf{Insance}} & \multicolumn{2}{c}{$MATLS$} & \multicolumn{2}{c}{$S5$} & \multicolumn{2}{c}{$CS2SA^*$} & \multicolumn{2}{c}{$RWS$} \\

\cmidrule(r){2-3} \cmidrule(r){4-5} \cmidrule(r){6-7} \cmidrule(r){8-9}

&  $Mean$      &  $RSD$  
&  $Mean$      &  $RSD$  
&  $Mean$      &  $RSD$  
&  $Mean$      &  $RSD$    \\

\toprule[0.1mm]

$eil76 $             &\textbf{22185}(1)                          & 0.75                        & 20097(3)                   & 0           & 18753(4)           & 0          & 21620(2)           & 0.02                   \\
$kroA100 $            &\textbf{42535}(1)                          & 1.45                    & 39440(3)                   & 0           & 39271(4)          & 0                & 41258(2)           & 0.3                   \\
$ch130$               &\textbf{61028}(1)                          & 0.12                    & 58685(2)                   & 1.21           & 50695(4)           & 0          & 57964(3)           & 4.6                   \\
$u159$                &58289(2)                          & 1.06                    & 57618(4)                   & 0           & 58090(3)          & 0          & \textbf{58946}(1)           & 1                   \\
$a280$              &\textbf{110132}(1)                          & 2.16                    & 109921(3)                   & 0           & 107696(4)           & 0          & 107874(2)           & 1.4                   \\
$u574$            &\textbf{254770}(1)                          &0.76                   & 251775(2)                   & 0.02          & 248584(3)          & 0           & 247992(4)           & 1.5                    \\
$u724$               &303435(4)                         & 1.17                    & 305977(2)                   & 0.32           & \textbf{309636}(1)           & 0          & 304420(3)           & 2.3                    \\
$dsj1000$                &340317(2)                          & 1.55                    & \textbf{342189}(1)                  & 0.59          & 332883(4)          & 0         & 339557(3)           & 0.68                    \\
$rl1304$              &572766(4)                         & 1.2                    & 575102(3)                   & 0.85          & 585600(2)          & 0         & \textbf{590103}(1)           & 0.02                   \\
$fl1577$            &609288(3)                          & 1.77                   & 607247(4)                   & 1.62           & \textbf{636422}(1)           & 0         & 635112(2)           & 0.1                    \\
$d2103$               &849625(2)                         & 1.35                   & \textbf{853587}(1)                   & 1.2           & 842520(4)          & 0          & 842596(3)           & 0.02                     \\
$pcb3038$                &1168108(4)                         & 0.52                    & 1179510(2)                 & 0.16          & \textbf{1193738}(1)          & 0         & 1176520(3)           & 0.55                     \\
$fnl4461$              &1617401(4)                         & 0.3                    & 1625856(2)                 & 0.16           & \textbf{1628414}(1)           & 0          & 1624685(3)           & 0.25                   \\
$pla7397$            &4178551(2)                          & 3.25                    & \textbf{4371433}(1)                   & 0.82           & 3713312(4)           & 0          & 3751665(3)           & 0.52                    \\
$rl11849$               &4587812(4)                          & 0.48                    & 4630753(3)                  & 0.29           & 4710135(2)           & 0          & \textbf{4729374}(1)           & 0.1                    \\
$usa13509$                &7767305(4)                          & 2.1                    & 7818115(3)                   & 0.86           & \textbf{8115168}(1)          & 0          & 8022398(2)           & 1.3                    \\
$brd14051$              &6492925(4)                          & 1.25                    & 6552658(3)                  & 0.58           & 6654162(2)          & 0          & \textbf{6778329}(1)           & 0.64                   \\
$d15112$            &6828152(4)                          & 2.3                   & 6991416(3)                  & 1.21           & \textbf{7606856}(1)           & 0          & 7596136(2)           & 0.41                    \\
$d18512$               &7164397(4)                          & 1.25                    & 7257669(3)                  & 0.81          & \textbf{7579996}(1)           & 0          & 7507146 (2)          & 0                     \\
$pla33810$                &15532942(4)                         & 1.5                    & 15574550(3)                   & 0.74           & 15756385(2)         & 0.8          & \textbf{15821323}(1)           & 0.3                     \\
\midrule

\multirow{1}{*}{\textbf{Average ranking}} & \multicolumn{2}{c}{$2.8$} & \multicolumn{2}{c}{$2.5$} & \multicolumn{2}{c}{$2.45$} & \multicolumn{2}{c}{$2.25$} \\



\bottomrule
\end{tabular}
\end{table}

As described in Section 3, We argue that the weight of the item has a greater impact on the final profit than other item attributes (value, location, etc.). In order to verify our speculation, the variant in equation \ref{eq:myScore} is configured as follows:
\begin{itemize}
    \item  Solver1: The value of exponent $\alpha$ is set to 1.5. 
    \item  Solver2: The value of exponent $\alpha$ is set to 1.

\end{itemize}
The results shown in Table \ref{Tab04} indicate that the Solver 2 performs better in many TTP instances. That is, this result verifies that our reasoning is plausible.

\begin{figure}
\centering
\begin{minipage}[c]{0.30\textwidth}
\centering
\includegraphics[height=3.5cm,width=4.2cm]{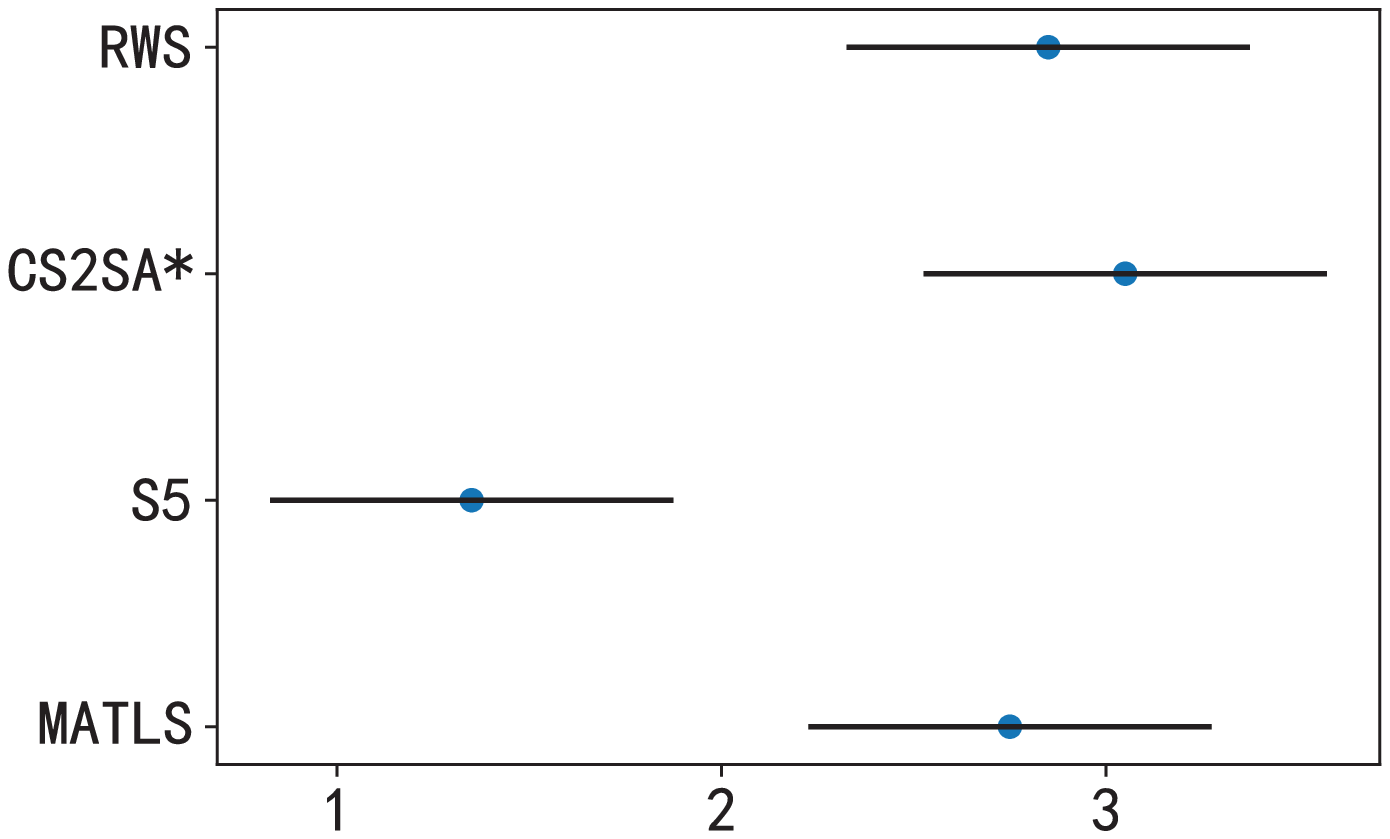}
\end{minipage}%
\begin{minipage}[c]{0.30\textwidth}
\centering
\includegraphics[height=3.5cm,width=4.2cm]{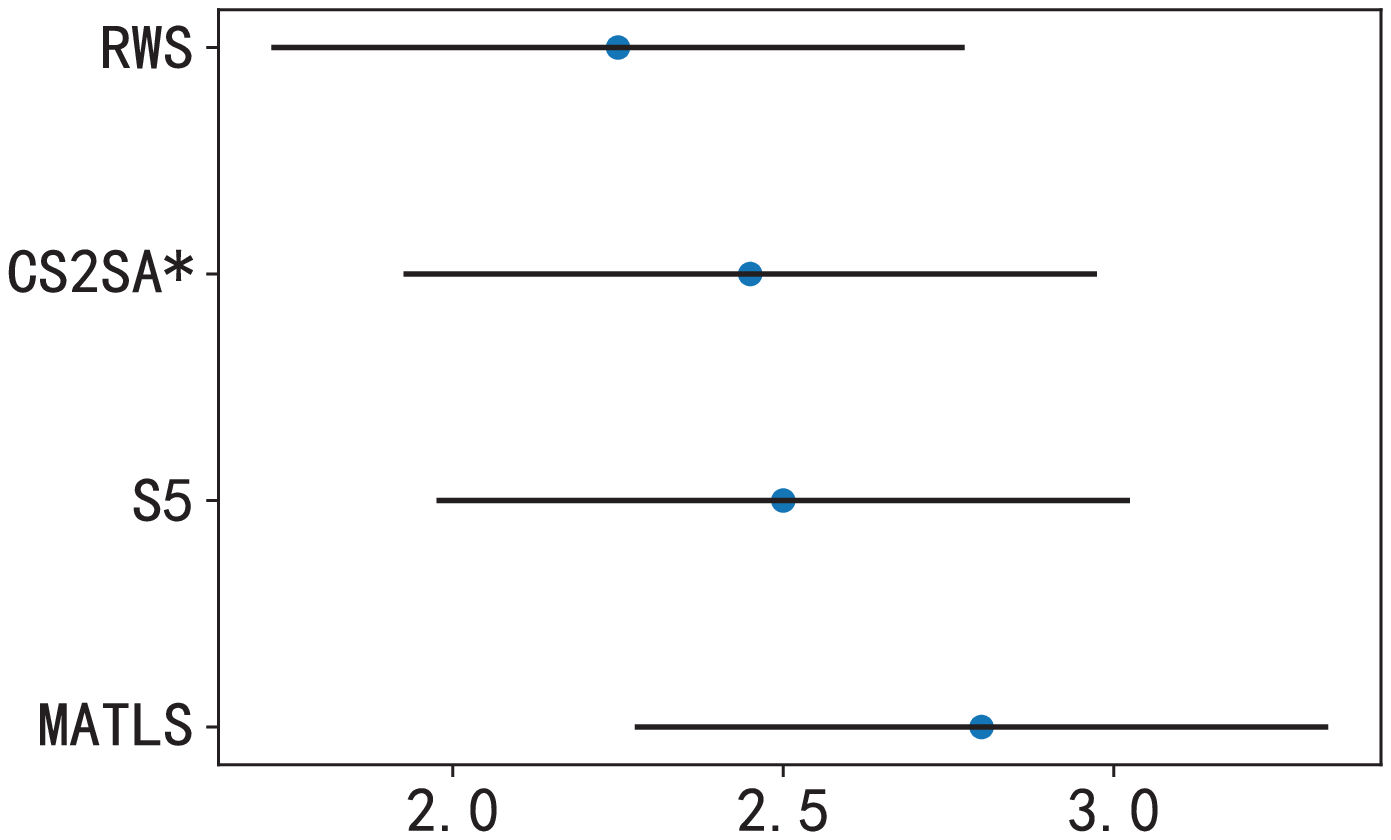}
\end{minipage}
\begin{minipage}[c]{0.30\textwidth}
\centering
\includegraphics[height=3.5cm,width=4.2cm]{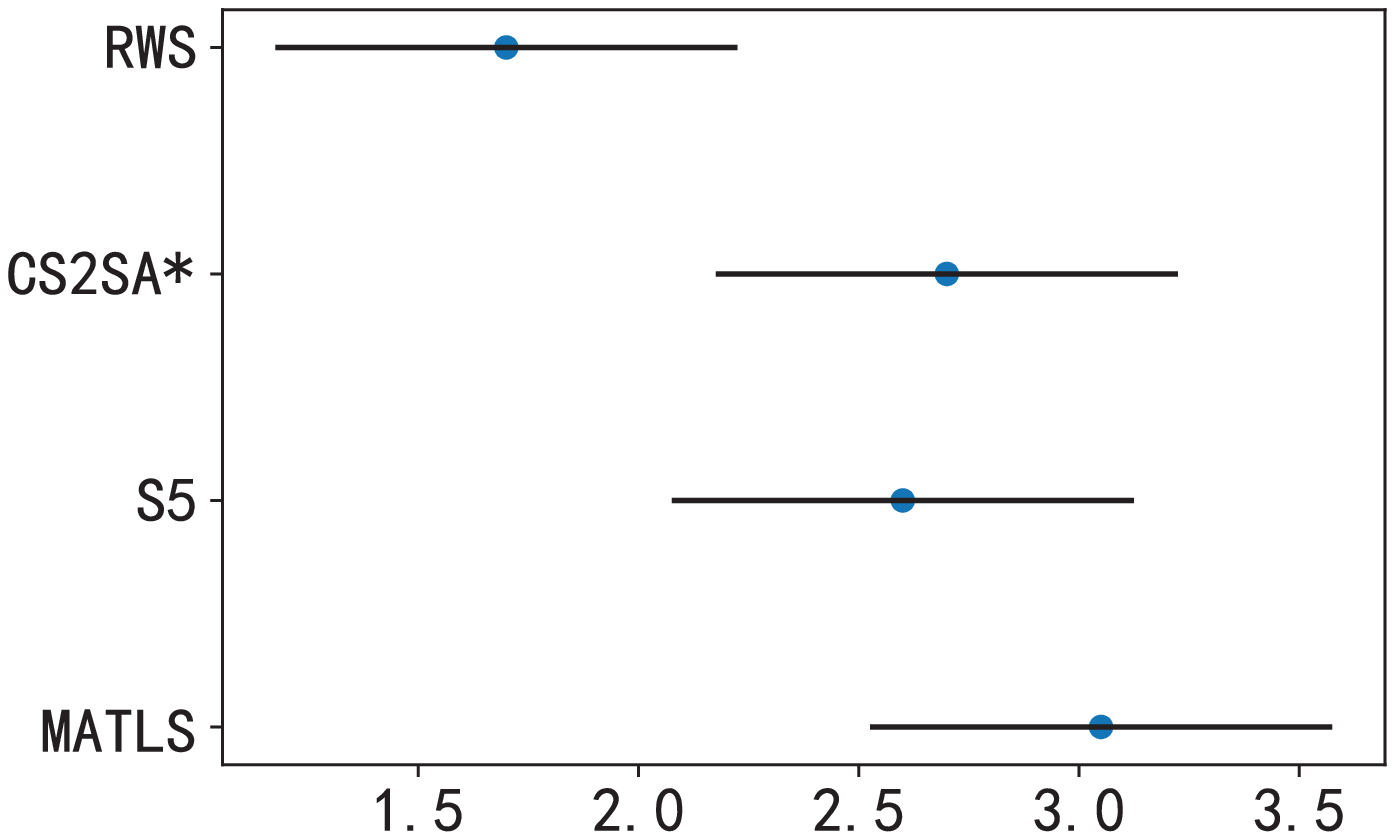}
\end{minipage}
\caption{Friedman test of four approaches in Category A, B, and C}
\label{fig:friedmanTest}
\end{figure}

\begin{figure}
\centering
\begin{minipage}[c]{0.50\textwidth}
\centering
\includegraphics[height=4.5cm,width=6.2cm]{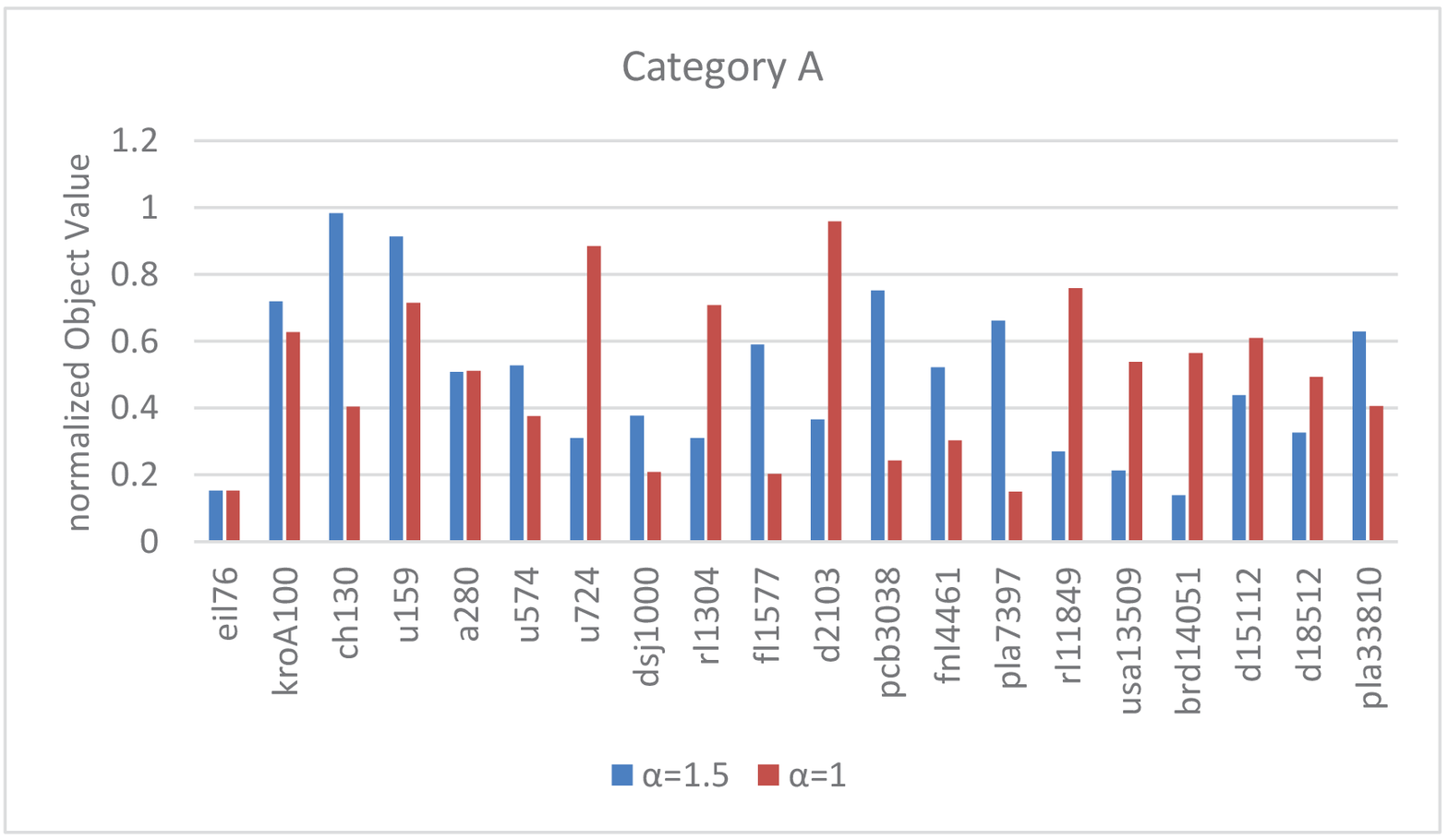}
\end{minipage}%
\begin{minipage}[c]{0.50\textwidth}
\centering
\includegraphics[height=4.5cm,width=6.2cm]{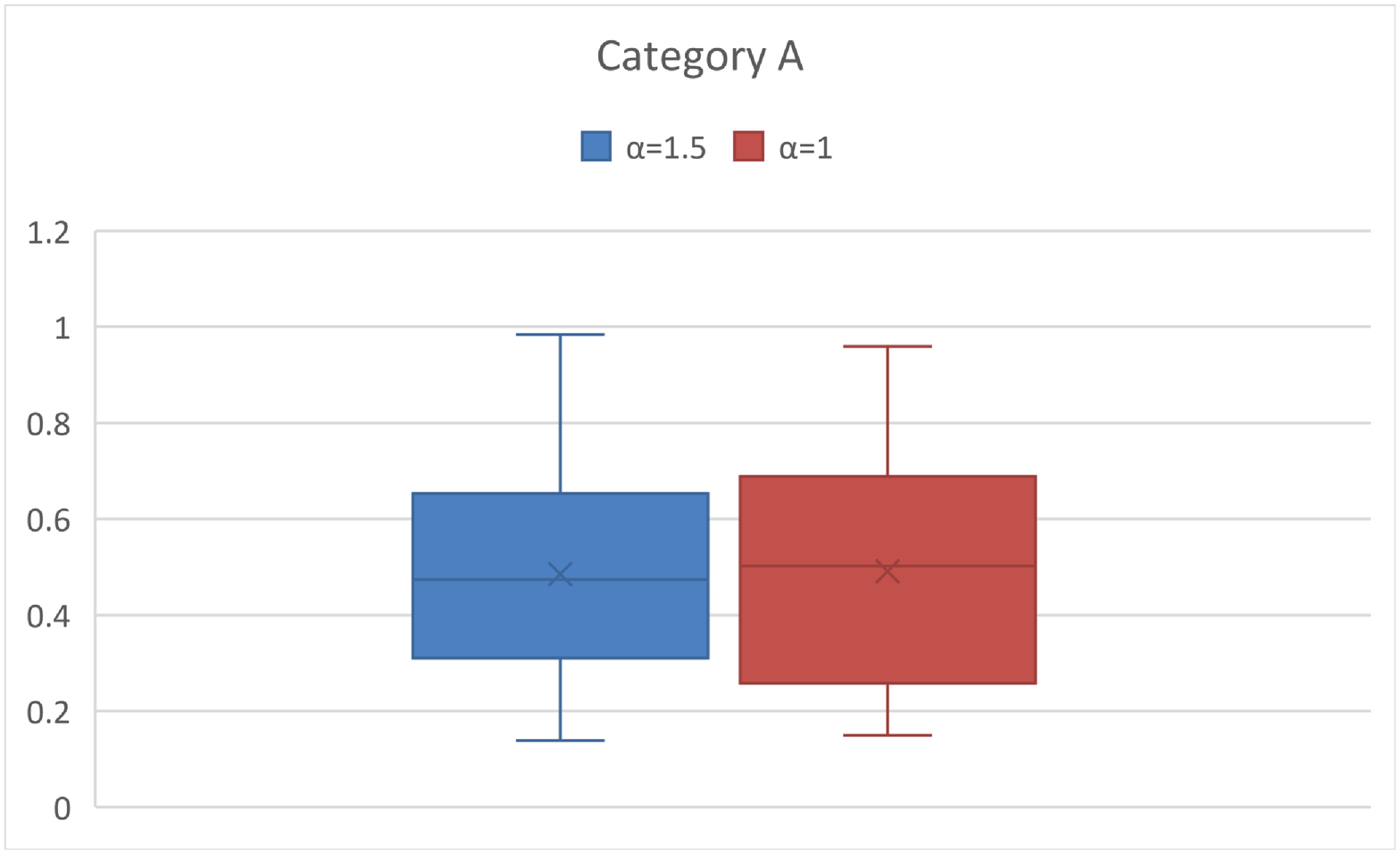}
\end{minipage}
\caption{Shown are the rescaled performances of our approaches with  parameter $\alpha$ on Category A instances}
\label{fig:barBoxA}
\end{figure}

\begin{figure}
\centering
\begin{minipage}[c]{0.50\textwidth}
\centering
\includegraphics[height=4.5cm,width=6.2cm]{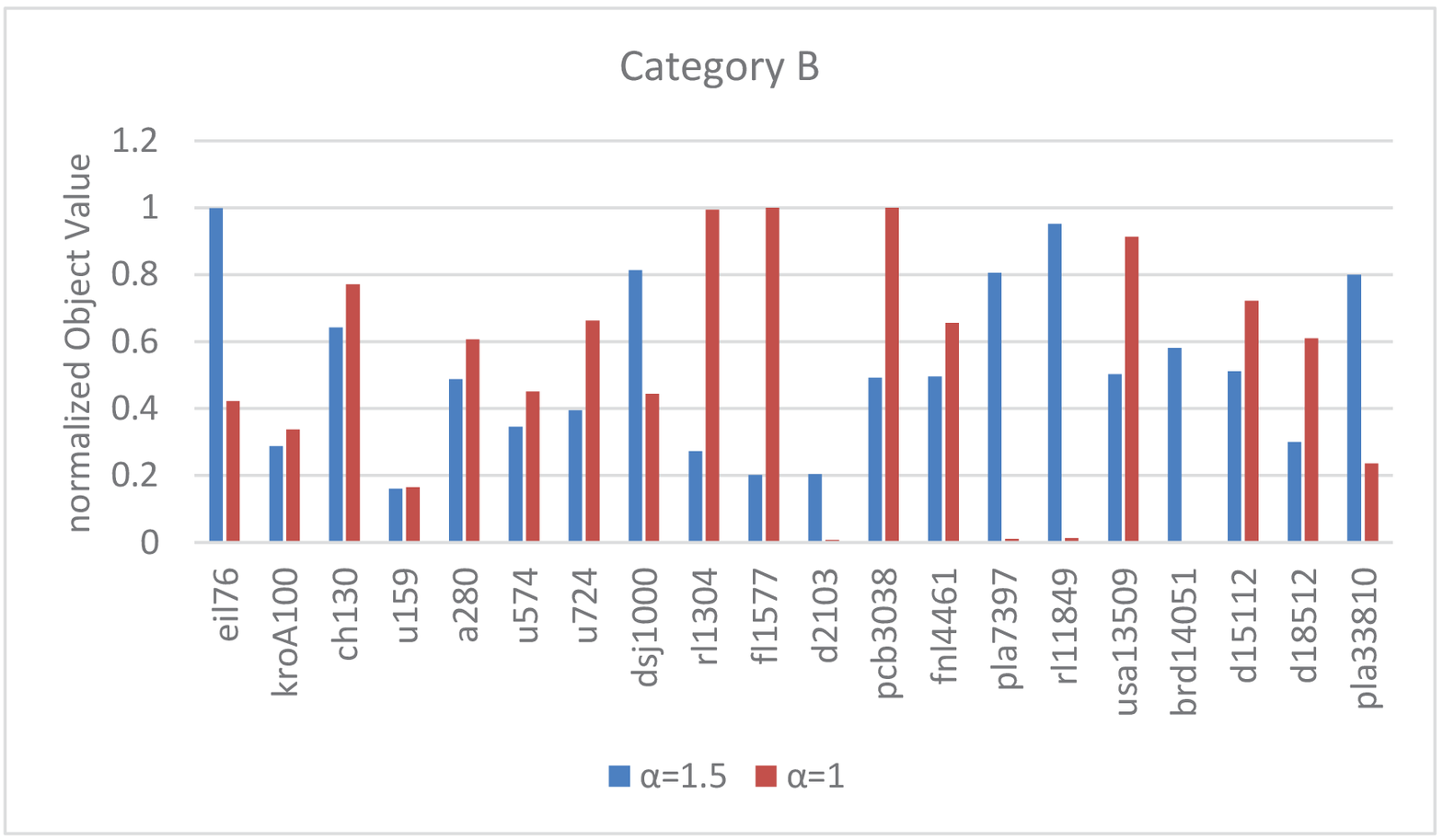}
\end{minipage}%
\begin{minipage}[c]{0.50\textwidth}
\centering
\includegraphics[height=4.5cm,width=6.2cm]{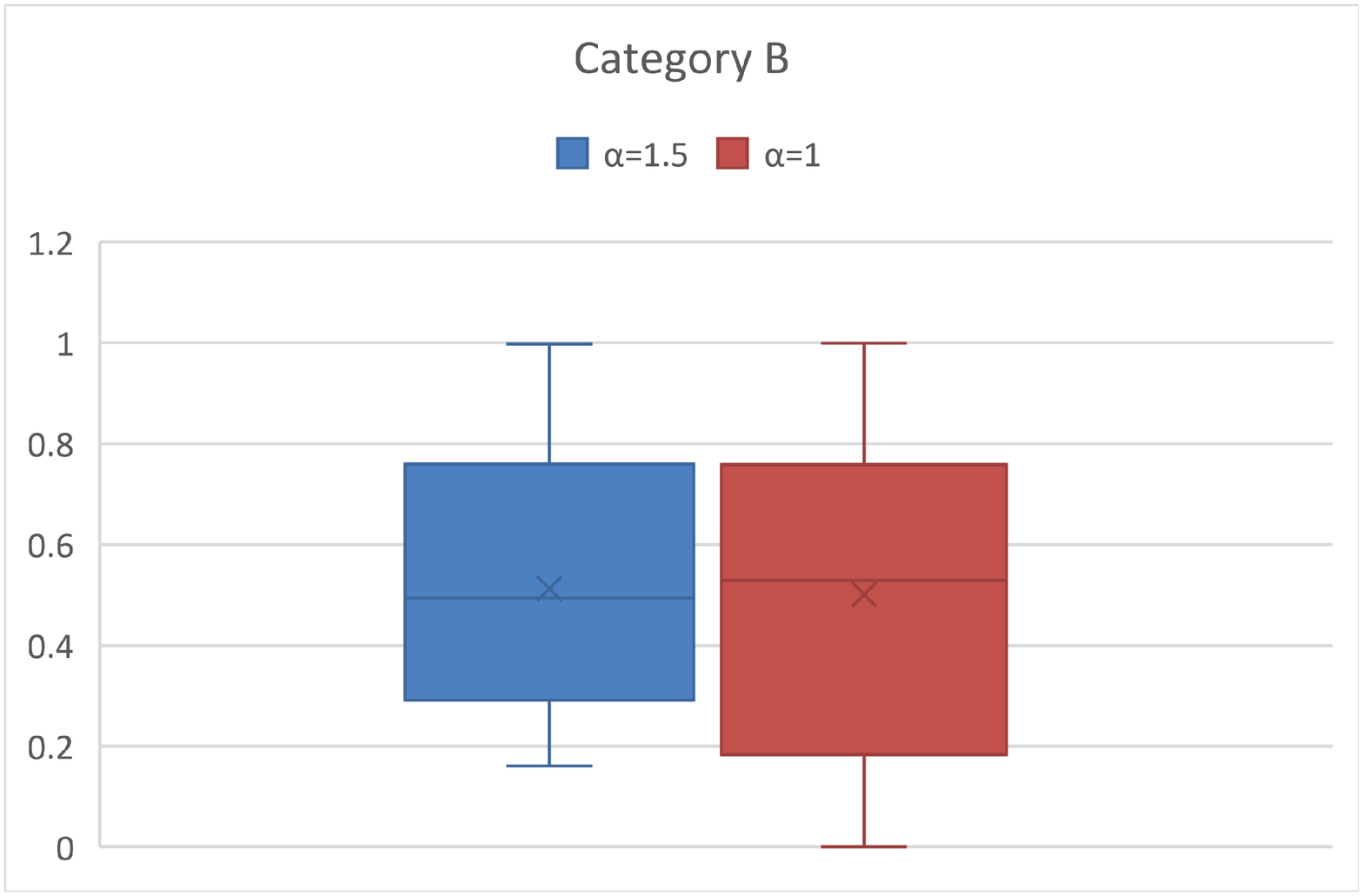}
\end{minipage}
\caption{Shown are the rescaled performances of our approaches with  parameter $\alpha$ on Category B instances}
\label{fig:barBoxB}
\end{figure}

\begin{figure}
\centering
\begin{minipage}[c]{0.50\textwidth}
\centering
\includegraphics[height=4.5cm,width=6.2cm]{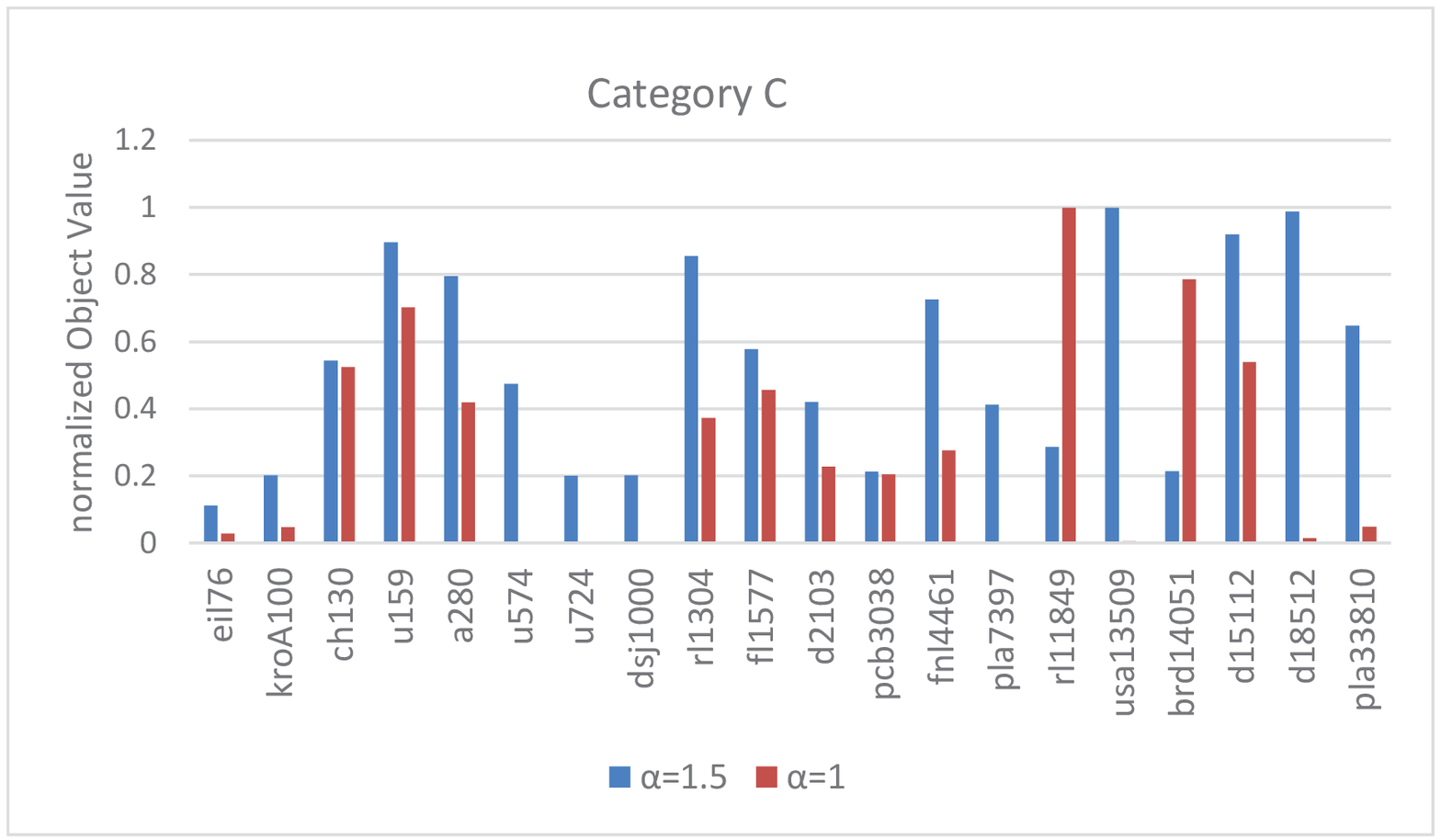}
\end{minipage}%
\begin{minipage}[c]{0.50\textwidth}
\centering
\includegraphics[height=4.5cm,width=6.2cm]{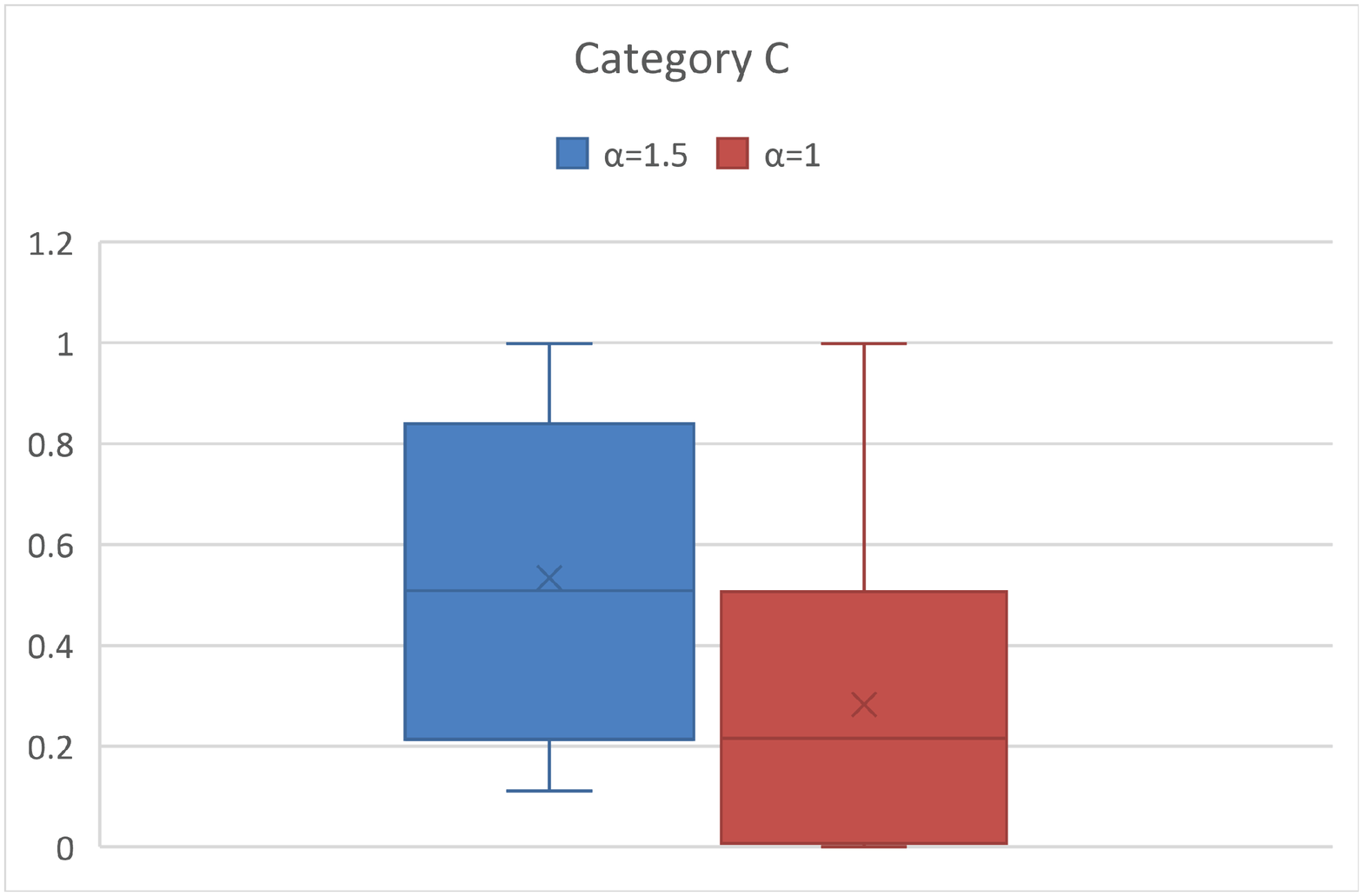}
\end{minipage}
\caption{Shown are the rescaled performances of our approaches with  parameter $\alpha$ on Category C instances}
\label{fig:barBoxC}
\end{figure}

\begin{figure}
\centering
\begin{minipage}[c]{0.50\textwidth}
\centering
\includegraphics[height=4.5cm,width=6.2cm]{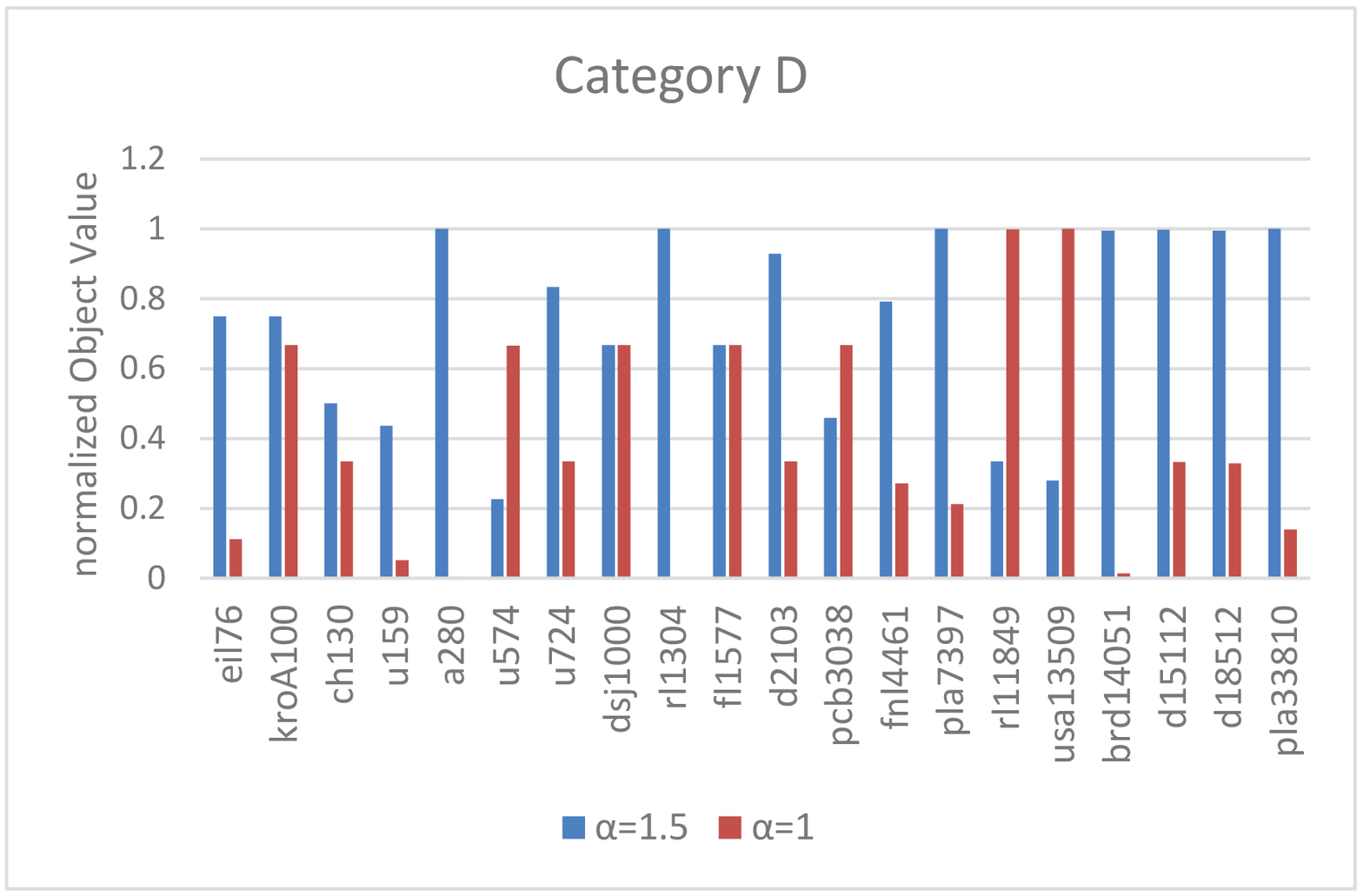}
\end{minipage}%
\begin{minipage}[c]{0.50\textwidth}
\centering
\includegraphics[height=4.5cm,width=6.2cm]{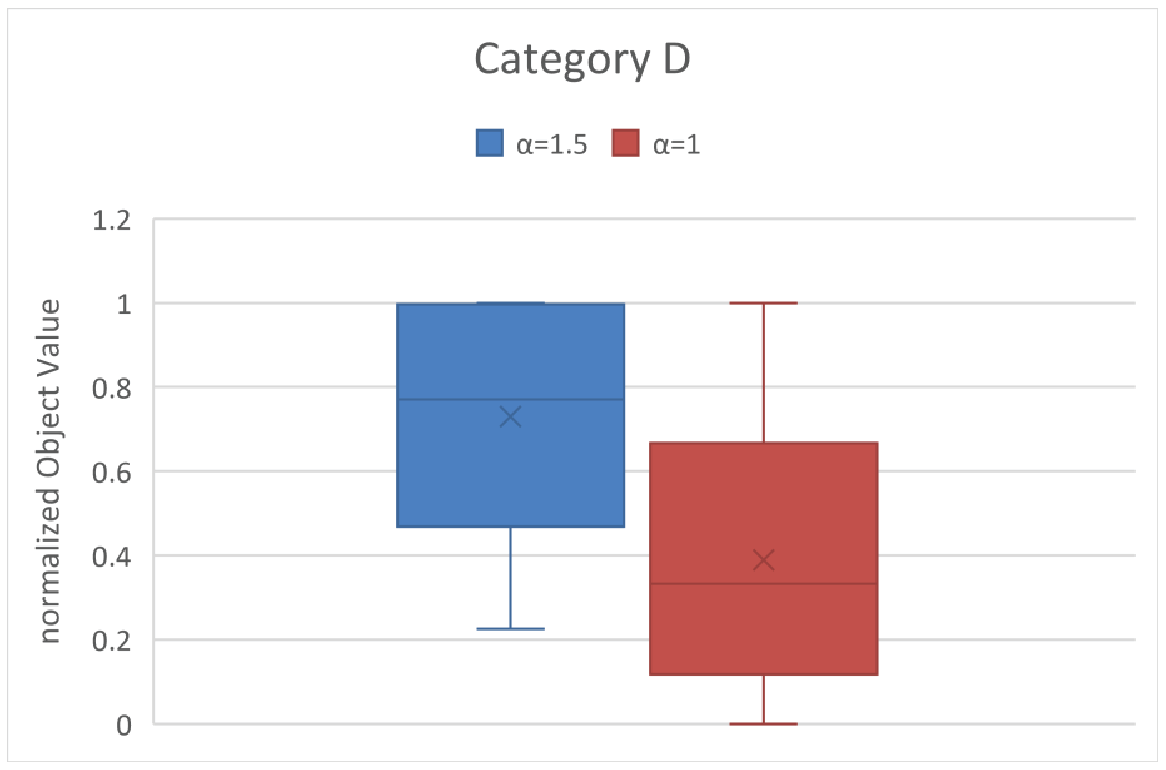}
\end{minipage}
\caption{Shown are the rescaled performances of our approaches with  parameter $\alpha$ on Category D instances}
\label{fig:barBoxD}
\end{figure}



\begin{table}[htbp]
\centering
\caption{Results for category C}
\label{Tab03}

\begin{tabular}{ccccccccc}

\toprule[0.3mm]

\multirow{2}{*}{\textbf{Insance}} & \multicolumn{2}{c}{$MATLS$} & \multicolumn{2}{c}{$S5$} & \multicolumn{2}{c}{$CS2SA^*$} & \multicolumn{2}{c}{$RWS$} \\

\cmidrule(r){2-3} \cmidrule(r){4-5} \cmidrule(r){6-7} \cmidrule(r){8-9}

&  $Mean$      &  $RSD$  
&  $Mean$      &  $RSD$  
&  $Mean$      &  $RSD$  
&  $Mean$      &  $RSD$    \\

\toprule[0.1mm]

$eil76 $             &\textbf{88115}(1)                          & 0.32                    & 85664(4)                   & 0          & 87577(3)          & 0            & 87664(2)           & 0.27                   \\
$kroA100 $            &155492(4)                          & 0.01                    & 155540(3)                   & 0           & 155585(2)          & 0         & \textbf{155947}(1)           & 0.48                   \\
$ch130$               &\textbf{203468}(1)                         & 2.13                    & 201085(3)                   & 0.82           & 197555(4)           & 0          & 202348(2)           & 1.9                   \\
$u159$                &242558(2)                          & 0.45                    & 242485(3)                  & 0.31           & 242201(4)          & 0.52          & \textbf{244770}(1)          & 0.05                   \\
$a280$              &426259(3)                         & 0.2                    & \textbf{429000}(1)                   & 0           & 421713(4)           & 0          & 426736(2)           & 0.14                   \\
$u574$            &966207(2)                          & 0.24                    & \textbf{966344}(1)                   & 0.11           & 953997(4)           & 0          & 955745(3)           & 0.16                    \\
$u724$               &1188761(3)                          & 0.45                    & 1188364(4)                   & 0.08           & 1191819(2)           & 0          & \textbf{1193604}(1)           & 0.32                    \\
$dsj1000$                &1472612(2)                          & 1.2                   & \textbf{1479605}(1)                   & 0.24           & 1468858(4)          & 0          & 1469206(3)           & 0.05                    \\
$rl1304$              &2178475(4)                          & 0.21                    & 2184853(3)                   & 0.33           & 2198943(2)           & 0          & \textbf{2198947}(1)           & 0                   \\
$fl1577$            &2466353(4)                          & 0.26                    & 2470917(3)                  & 0.21           & 2505291(2)          & 0          & \textbf{2505295}(1)           & 0                    \\
$d2103$               &3392866(2)                          & 0.32                    & 3392172(3)                   & 0.26           & 3373781(4)          & 0          & \textbf{3410978}(1)           & 0.93                     \\
$pcb3038$                &4564228(4)                          & 0.22                    & 4573748(3)                   & 0.15           & 4612956(2)          & 0          & \textbf{4612966}(1)           & 0                     \\
$fnl4461$              &6534422(4)                          & 0.17                    & \textbf{6554497}(1)                   & 0.26           & 6545335(3)           & 0          & 6545355(2)           & 0                   \\
$pla7397$            &13865791(2)                          & 1.55                    & \textbf{14239606}(1)                   & 1.2           & 13197756(4)           & 0          & 13440471(3)           & 1.83                    \\
$rl11849$               &18275210(4)                          & 0.23                    & 18314650(3)                   & 0.12           & \textbf{18505203}(1)           & 0          & 18422410(2)           & 0.09                    \\
$usa13509$                &25878184(4)                          & 0.44                    & 25918971(3)                  & 0.55           & 26437361(2)          & 0          & \textbf{26552971}(1)           & 0                    \\
$brd14051$              &23672405(4)                          & 0.62                    & 23826398(2)                   & 0.51           & \textbf{23908540}(1)          & 0          & 23809751(3)           & 0.01                   \\
$d15112$            &25942410(4)                          & 1.52                    & 26211252(3)                   & 1.04           & 27182609(2)           & 0.15          & \textbf{27184251}(1)           & 0                    \\
$d18512$               &27164388(4)                          & 1.25                    & 27427144(3)                   & 0.32           & 27849746(2)           & 0.21          & \textbf{27980876}(1)          & 0                     \\
$pla33810$                &58003895(3)                          & 0.5                   & 57967586(4)                  & 0.42           & 58107703(2)         & 0.01          & \textbf{58818293}(1)           & 0.21                    \\
\midrule

\multirow{1}{*}{\textbf{Average ranking}} & \multicolumn{2}{c}{$3.05$} & \multicolumn{2}{c}{$2.6$} & \multicolumn{2}{c}{$2.7$} & \multicolumn{2}{c}{$1.7$} \\



\bottomrule
\end{tabular}
\end{table}

\begin{table}[htbp]
\centering

\caption{Performance comparision of two solvers on 3 categories of TTP instances }

\label{Tab04}

\begin{tabular}{cccccccc}

\toprule[0.3mm]

\multirow{2}{*}{\textbf{Insance}} & \multicolumn{2}{c}{$Category\ A$} & \multicolumn{2}{c}{$Category\ B$} & \multicolumn{2}{c}{$Category\ C$}  \\

\cmidrule(r){2-3} \cmidrule(r){4-5} \cmidrule(r){6-7} 

&  $Solver 1$      &  $Solver 2$  
&  $Solver 1$      &  $Solver 2$   
&  $Solver 1$      &  $Solver 2$      \\

\toprule[0.1mm]

$eil76 $             &\textbf{3765}                          & 3670                    & \textbf{21620}                   & 20192           & \textbf{87664}          & 87599                            \\
$kroA100 $            &\textbf{4445}                          & 4424                    & 41258                   & \textbf{41353}           & \textbf{155947}           & 155669                          \\
$ch130$               &\textbf{9013}                         & 8963                    & 57964                   &\textbf{ 58792}           & \textbf{202348}           & 202182                           \\
$u159$                &\textbf{8627}                          & 8566                    & \textbf{58966}                   & 58955           & \textbf{244770}          & 244228                            \\
$a280$              &17723                          & 17723                    & 107874                  & \textbf{108378}           & \textbf{426736}           & 424358                             \\
$u574$            &\textbf{26366}                          & 26265                    & 247992                  & \textbf{249368}          &\textbf{955745}           & 953998                          \\
$u724$               &48794                          & \textbf{49588}                    & 304420                  & \textbf{309750}           & \textbf{1193604}           & 1191819                           \\
$dsj1000$                &\textbf{141117}                         & 140620                   & \textbf{339557}                   & 338661           & \textbf{1469206 }         & 1468859                           \\
$rl1304$              &75206                          & \textbf{76435}                    & 585103                   & \textbf{585600}           & \textbf{2198947}           & 2198942                        \\
$fl1577$            &\textbf{88923}                          & 88248                   & 635112                   & \textbf{636424}           & \textbf{2505295}           & 2505294                            \\
$d2103$               &118338                         & \textbf{118652}                    & \textbf{842596}                   & 842522           & \textbf{3410978}           & 3393849                              \\
$pcb3038$                &\textbf{149337}                          &146115                    & 1176520                   & \textbf{1193737}           & \textbf{4612966}          & 4612957                             \\
$fnl4461$              &\textbf{241291}                          & 240822                    & 1624685                   & \textbf{1628417}           & \textbf{6545355}           & 6545346                            \\
$pla7397$            &\textbf{315386}                         &314073                    & \textbf{3751665}                   & 3713312           & \textbf{13440471}          & 13197751                           \\
$rl11849$               &653857                         &\textbf{659283}                    & \textbf{4729274}                   & 4710149           & 18422410           & \textbf{18504597}                           \\
$usa13509$                &677983                          & \textbf{682238}                    & 8022398                   & \textbf{8115207}           & \textbf{26552971}          & 26422272                      \\
$brd14051$              &798787                         &\textbf{802188}                    & \textbf{6778329}                   & 6654177           & 23809751           & \textbf{23907953}                       \\
$d15112$            &868019                         &\textbf{868998}                    & 7606136                   & \textbf{7606876}           & \textbf{27184251}           & 27182054                          \\
$d18512$               &962781                          &\textbf{964518}                    & 7507146                   & \textbf{7580272}           & \textbf{27980877}           & 27861162                           \\
$pla33810$                &\textbf{1781384}                          & 1777592                    & \textbf{15821323}                  & 15745060           & \textbf{58818293}              & 58545275                     \\


\bottomrule
\end{tabular}
\end{table}

\subsection{Results analysis and discussion}

According to the presented results, the proposed algorithm surpasses the other state-of-the-art algorithms (MATLS\cite{RN20}, S5\cite{RN19}, and CS2SA*\cite{RN9}) for most instances of the TTP. This is mainly due to the solution space is explored more adequate. The simple Bit-Flip heuristics and the simulated annealing algorithm are commonly used for KP component of TTP. However, we run some instances and the result
shows that the simulated annealing algorithm has better performance on the large-scale instances. The proposed algorithm adopts a reverse order picking plan, based on sorting the items according to the item's profits, weights, and location in the given tour, and picks the item with a score greater than the average. In order to avoid falling into a local optimal situation, the algorithm uses different travel tour instead of fixed one in a given time budget.
The algorithm has some competitiveness in category A even the profit and the weight of the items are strongly correlated as shown in Table \ref{Tab01}. The presented results show that S5 surpasses the other algorithms in the most instances. The category A has the smallest knapsack capacity and only one item in each city. We argue the greedy approach adopted by S5 is beneficial in solving this type of KP component of the TTP.\par

For the instances with a higher knapsack capacity in category B (5 items in each city, KP uncorrelated with similar weights), the comparative results suggeste that  the CS2SA* and S5 still competitive in this type of the category. However, Table \ref{Tab02} show that RWS clearly outperforms the other heuristics for the majority of the instances  such as u159, rl1304, rl11849, brd14051, pla33810. MATLS and CS2SA*
also perform better in some instances which are shown in the table. \par

 Table \ref{Tab03} shows the comparative results for Category C (10 items per city, uncorrelated). This category has the largest knapsack capacity. The CS2SA* perform better in many instances compared to other algorithms shown in the table. Note that the RWS outperfoms the other heuristics in the most instances for high knapsack capacities such as kroA100, u159, u724, rl1304, fl1577, d2103, pcb3038, usa13509, d15112, d18512, pca33810. However, the MATLS perform poorly, it is mainly because the population-based heuristics for the TTP  is not efficient for handling large-scale instances. To get a better performance analysis of all algorithms we perform,
The Friedman’s test is applied to find the differences between
the groups when the dependent variable is ordinal. The Nemenyi post-hoc test after the Friedman test is applied .The test ranking of all the algorithms is shown in Figure \ref{fig:friedmanTest}, the friedman test ranking of tour approaches in category A,B and C are presented. As can be seen from the above figure, in large-scale instances (in the Category C), our proposed algorithm ranks perform better the other algorithms.
\par

In addition, to verify our suppose mentioned in section 3, Due to the cumulative effect of the weight of the picked item, the weight of the item has a greater impact on the final profit than other item attributes (value, location, etc.). We did some experiments and the results are shown in  table \ref{Tab04}. The value of exponent $\alpha$ isused to manage the impact of the weight of items on the final profits.
It can be clearly observed that the Solver1 outperforms the Solver2 in the many instances (especially in the Category C). A representative excerpt of the results  is shown in Figure \ref{fig:barBoxA}, \ref{fig:barBoxB}, \ref{fig:barBoxC}. Note that we rescale the achieved objectives values into the range $\left[ 0, 1 \right]$ by normalization method. The box diagram on the right side of the picture shows the performance of the algorithm with the different parameter $\alpha$. The results shown that in small-scale (Category A) and medium-scale (Category B) instances, the performance of the solver1 algorithm has no obvious superiority. However, 
in large-scale instances (Category C), the performance of the solver1 algorithm has greater advantages. For further verification, we compared two large-scale instances of experiments in Category C (item factor: 10) and D (item factor: 9) in Figure \ref{fig:barBoxD}. This experimental result also proves the superior performance of solver1. In other words, the result of this investigation verifies our suppose: the weight of the item has a greater impact on the final profit in the large-scale instances.\par

Therefore, from the result, we can conclude that our proposed algorithm performs better than the other state-of-the-art algorithms on most of the instances, espically for category B and category C. The proposed algorithm adopts a reverse order picking plan, based on sorting the items according to the proposed formula. It expands the search space and is likely to find potential solutions for solving the TTP.

\section{Conclusion}

In real-world optimisation problems, combinatorial optimization problems with two or more interpendent components have a major role. Due to the interpendency, an optimal solution to one of components does not gurantee the overall solution to the whole problem. The TTP can be thought of as a combination of two interdependent well-known problems namly the Travelling Saleman Problem (TSP) and 0-1 Knapsack Problem (KP) which was introduced to represent the real-world applications. The interaction and dependence between the sub-problems indicate the complexity of the whole problem. Some approaches have been introduced to solve this problem, such as heuristic, cooperative methods and other methods etc. \par

In this paper, due to the cumulative effect of the weight of the picked item, we suppose that the weight of the item has a greater impact on the final profit than other item attributes (value, location, etc.) , To address the issue, we proposed a new heuristic for the TTP based on managing the impact of the weight of items on the final profits. Besides, we believe that high-value and low-weight items near the end of the travel route should be picked up, Under the condition that the total picked-up item weight is not weightier than the knapsack capacity, the items will be picked up from back to front according to the route, So we proposed a method of picking items in reverse order. The obtained results show that our approach are competitive for many instances of different sizes and types compared other heuristics. Especially, our algorithm performs better in large-scale instances.
\par

Most real-world combinatiorial optimization problems have more than two components. In the future, further research will be made to investigate other problems with more than two components to get the internal dependencies.  Furthermore, our proposed method can be further improved in sapce exploration and adopted in problems with many interacting component that have great potential in the real-world applications.

\section *{Acknowledgments}
This work was partially supported by the Natural Science Foundation of Guangdong Province of China (Grant No.2020A1515010691), Science and Technology Project of Guangdong Province of China (Grant No.2018A0124), and National Natural Science Foundation of China (Grant Nos. 61573157 and 61703170). The authors also gratefully acknowledge the reviewers for their helpful comments and suggestions that helped to improve the presentation.

\bibliographystyle{spmpsci}
\bibliography{refs}